\documentclass[lettersize,journal]{IEEEtran}
\usepackage{amsmath,amsfonts}
\usepackage{algorithmic}
\usepackage{algorithm}
\usepackage{array}
\usepackage[caption=false,font=scriptsize,labelfont=footnotesize,textfont=sf]{subfig}
\usepackage{textcomp}
\usepackage{stfloats}
\usepackage{url}
\usepackage{verbatim}
\usepackage{graphicx}
\usepackage{cite}
\usepackage{multirow}
\usepackage{booktabs}
\usepackage{longtable,siunitx}

\begin{document}

\title{Comprehensive and Reliable Feature Attribution for Diverse Modalities and Models via Frequency-Domain Insights}


\author{
\thanks{
    This work was supported by the National Natural Science Foundation of China No. 62372340, the Major Technical Research Project of Hubei Province No. 2023BAA018, and Key Research and Development Program Projects of the Corps Nos. 2024AA010, 2024AB080, 2025AB052, Xinjiang Production and Construction Corps Key Laboratory of Computing Intelligence and Network Information Security Open Fund No. CZ002702-02. \emph{(Zechen Liu and Feiyang Zhang contributed equally to this work.) (Corresponding Author: Wei Song.)}}
    Zechen Liu\thanks{Zechen Liu is with the School of Computational Science, Wuhan University, Wuhan, China (e-mail: zecliu@whu.edu.cn).},
    Feiyang Zhang\thanks{Feiyang Zhang is with the Brain Research Center, Wuhan University, Wuhan, China (e-mail: fyzhang.neu@whu.edu.cn).},
    Wei Song\thanks{Wei Song is with the School of Computational Science, Wuhan University, Wuhan, China, and also with the College of Information Science and Technology (School of Cyber Science and Technology), Shihezi University, Shihezi, China (e-mail: songwei@whu.edu.cn).},
    Yuqi Zhang\thanks{
    Yuqi Zhang is with the College of Information Science and Technology (School of Cyber Science and Technology), Shihezi University, Shihezi, China (e-mail: zhangyuqi@stu.shzu.edu.cn).
    },
    Wei Wei\thanks{Wei Wei is with the Brain Research Center, Wuhan University, Wuhan, China (e-mail: wei.wei@whu.edu.cn).},
    Xiang Li\thanks{Xiang Li is with the Brain Research Center, Wuhan University, Wuhan, China (e-mail: li.xiang@whu.edu.cn).}
}

\markboth{Journal of \LaTeX\ Class Files,~Vol.~14, No.~8, August~2021}%
{Liu \MakeLowercase{\textit{et al.}}: A Sample Article Using IEEEtran.cls for IEEE Journals}

\maketitle



\begin{abstract}
Interpreting deep neural networks remains challenging because spatial features---the dominant basis used in existing attribution methods---lack a mathematically valid null baseline. This makes systematic ablation ill-posed and undermines the faithfulness and reliability of spatial-domain explanations. In this work, we established frequency components as a universal and theoretically grounded alternative for attribution. Unlike spatial features, frequency components inherently admit a well-defined zero-power baseline, enabling principled and reliable ablation.
We further introduce a computationally efficient frequency-domain attribution framework based on control theory. Comprehensive evaluations on ImageNet and CIFAR-10 demonstrate the superior faithfulness of our approach compared to spatial-domain baselines. Extensive experiments across diverse modalities---including text, medical images, and audio---and model architectures---such as ResNet, Vision Transformers, U-Net, and large language models---confirm the stability and general applicability of our framework.
An important empirical observation is that models, regardless of the data modality, base their predictions on a remarkably small set of frequency features. For Vision Transformers, 4\% of features suffice for 80\% of predictions. These critical components exhibit markedly stronger structural regularities than spatial features. Notably, reconstructing only these components back into the spatial domain reveals detailed and semantically meaningful patterns that traditional saliency-based attribution methods fail to capture. Our code is available at https://github.com/Zechen6/FFC-FastFourierCorrelation.

\end{abstract}


\section{Introduction}
\begin{figure}
    \centering
    \includegraphics[width=\linewidth]{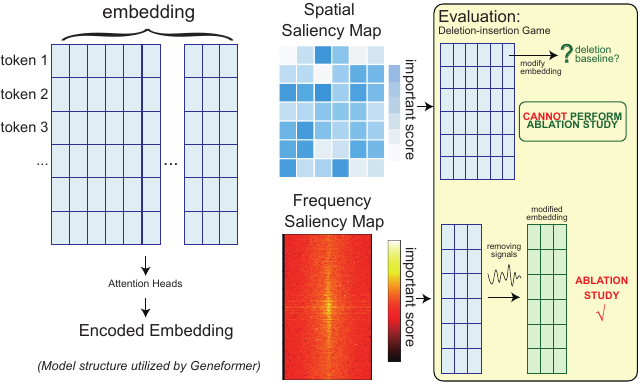}
    \caption{Spatial features cannot be subjected to principled ablation because no well-defined non-effect baseline exists.
In contrast, Fourier features admit a clear nullification operation: one can simply set their spectral power to zero.}
    \label{fig:intro-a}
\end{figure}

Feature attribution methods are one of the most widely used post-hoc explanation methods for interpreting the decision basis of deep neural networks. However, their explanations are seriously criticized by numerous researchers~\cite{unstable,unstableAAAI,unstablenips,unstablenips2020,unstabletheo22,unstabletheo23,Quanshi,LPI,attbenchmark,aggattri,conflict1,conflict2} for their unreliability. The primary challenge lies in the absence of a well-defined non-effect baseline for spatial features. While ablation studies---which remove features to observe changes in model output---are widely accepted for evaluating attribution faithfulness, traditional methods struggle to implement them effectively due to this missing baseline, ultimately undermining the trustworthiness of their results. As shown in Fig.~\ref{fig:intro-a}, determining a non-effect baseline for an individual embedding value is problematic: no matter which baseline we choose, the feature still consists of the vector. This phenomenon also exists in both images and audio.

\begin{figure*}[t]
    \centering
    \subfloat[]{
    \includegraphics[width=0.8\linewidth]{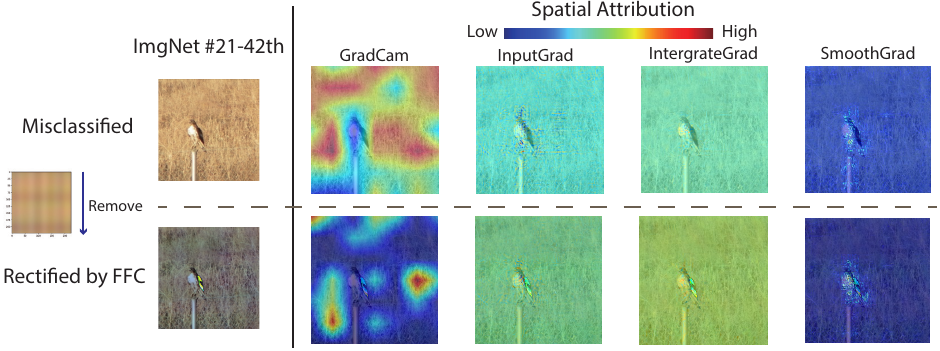}
    \label{fig:intro-b}
    }\\
    \subfloat[]{
    \includegraphics[width=0.8\linewidth]{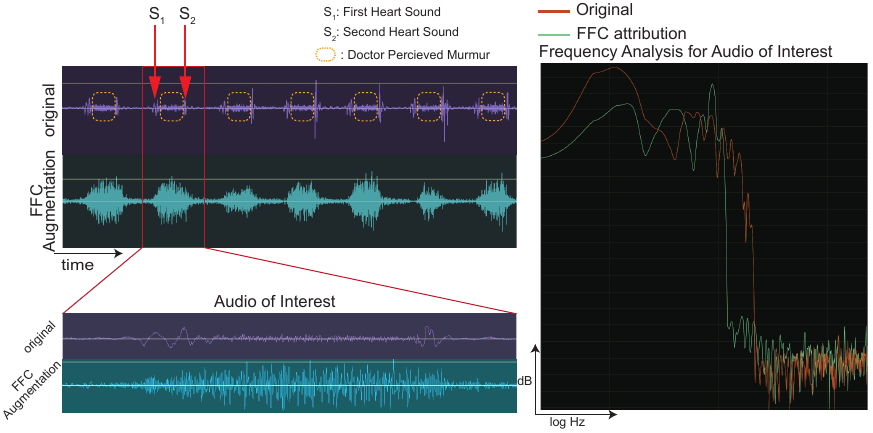}
    \label{fig:intro-c}
    }
    \caption{(a) Fourier features can provide far deeper insights than those offered by saliency maps. In this example, the sample was originally misclassified by a Vision Transformer. After removing the top 15 Fourier features identified by FFC, the network correctly classifies the sample. Interestingly, Smooth-Grad and Grad-CAM produce divergent explanations, revealing the instability inherent to spatial attribution. (b) In audio tasks, the limitations of saliency maps become particularly evident. While saliency maps can only highlight regions of interest, Fourier features provide more detailed information about which frequency components influence the network's decisions. For example, we trained a ResNet-50 for a cardiac auscultation task. Although the ground truth of the heart murmur is indicated by a physician, saliency maps can merely highlight the relevant region. In contrast, FFC can amplify the signals within the target area, offering more precise and informative guidance than traditional spatial attribution methods. As shown in the right panel, FFC enables a frequency-domain analysis that clearly identifies which components are amplified.}
    \label{fig:intro-more-insight}
\end{figure*}

To overcome these challenges, we shift our perspective to the frequency domain, where the aforementioned issues are fundamentally resolved. A frequency component—termed a Fourier feature in this work---is rooted in well-established mathematical and signal processing theory. Crucially, each Fourier feature inherently admits a clear non-effect baseline: its power can be precisely set to zero. Thus an ablation study can be reliably implemented to validate its importance. Surprisingly, as shown in Fig.~\ref{fig:intro-more-insight}, Fourier features provide not only reliable results but also deeper insights into the decision-making process, beyond what is conveyed by the saliency map. This is because Fourier features encapsulate phase, frequency, and magnitude information, whereas spatial features only represent magnitude. 

However, constructing a frequency-domain attribution method is far from straightforward. The existing foundations of real-valued attribution do not directly carry over to the complex domain. While applying Shapley-based~\cite{sharpe1966mutual} attribution is conceptually simple, its computational cost becomes prohibitive in this setting---requiring, in some cases, years to interpret even a single dataset.

To bridge this gap, we propose Fast Fourier Correlation (FFC)---a computationally efficient and practical frequency-domain attribution method, inspired by principles of feedback control and signal correlation. We first modify the input sample by performing gradient descent while keeping the network parameters fixed. The resulting modified sample is treated as the ``expected signal” that minimizes the network loss. We then compute the correlation between the expected signal and the original input. The frequency components preserved in the expected signal are regarded as important, as they represent the features the network relies on most.

Our comparative experiments on benchmark datasets demonstrate that FFC outperforms its spatial counterparts in terms of faithfulness, as evidenced by the deletion-insertion game---an established evaluation metric based on the ablation paradigm. Notably, FFC is the only method that increases model confidence after removing low-score features, thereby confirming that these low-score features indeed interfere with the model's decision-making, while the remaining features play a crucial role. 
We then perform maintain-rate experiments to validate the importance of the high-score features we identified. The results reveal that even when $90\%$ of the low-score features are deleted, the network is still able to maintain its top-one prediction for most samples, further reinforcing the faithfulness of FFC. 
Extensive experiments across a range of models and modalities show that our faithfulness remains consistently high in text, audio, and MRI. Notably, in the MRI segmentation task, the model itself already highlights the important regions, making the saliency map (commonly generated by traditional methods) redundant. 

Surprisingly, our experimental results uncover that, although the tested data appear highly unstructured in the spatial domain, their structured characteristics are markedly more pronounced in the frequency domain. Specifically, we observe that: (1) only a small and more fixed subset of Fourier features is sufficient to support network classification---for instance, Vision Transformers (ViT) classifying ImageNet require merely 4\% of Fourier features to correctly classify 80\% of samples; (2) the contributing Fourier features are more spatially concentrated, unlike spatial features whose positions are highly uncertain; and (3) the features exhibit high specificity, making Fourier features more effective for classification. The comparison between spatial and Fourier features is displayed in Table ~\ref{tab:comparison}.

\begin{table}[]
    \centering
\caption{Fourier Features and Spatial Features}
\label{tab:comparison}
\begin{tabular}{@{}p{3.5cm}ccc@{}}
\toprule
 & Fourier Feature & Spatial Feature  \\
\midrule
Order and Position & \textbf{Fixed}$\uparrow$ & Arbitrary$\downarrow$ \\
Semantic Meanings & \textbf{Higher}$\uparrow$ & Lower$\downarrow$ \\
Information Concentration & \textbf{Higher}$\uparrow$  & Lower$\downarrow$ \\
Number of Features Required for Classification & \textbf{Lower}$\downarrow$ & Higher$\uparrow$ \\
\bottomrule
\end{tabular}
\end{table}

Our main contributions are:
\begin{itemize}
    \item{\textbf{We uncover a new perspective (the frequency domain) to overcome the unreliability} of traditional attribution methods, and for the first time, establish the theoretical validity of rigorous ablation analysis within this domain.}
    \item{\textbf{We propose the first universal frequency-domain attribution method} that is applicable across modalities, maintains low computational overhead, and achieves higher faithfulness than all existing spatial-based approaches.}
    \item{\textbf{We reveal new structural properties of deep models in the frequency domain}, showing that high-contributing components are sparse, concentrated, and semantically aligned across samples.}
\end{itemize}

\section{Related Works}
\subsection{Studies of Spatial Attribution Methods and Metrics}
Spatial attribution methods have received substantial attention due to their ability to produce intuitive heatmaps that align with human perception~\cite{inputgrad}. A wide range of techniques has been proposed, rooted in different theoretical principles. Integrated Gradient-based approaches include Integrated Gradients (IG)~\cite{intgrad}, LPI~\cite{LPI}, MIG~\cite{MIG}, and Expected Gradients (EG)~\cite{EG}, while perturbation-based methods such as Smooth-Grad~\cite{smoothgrad} and LIME~\cite{LIME} rely on systematically modifying inputs. Backpropagation-based approaches, including DeepLIFT~\cite{deeplift}, Input×Gradient~\cite{inputgrad}, Guided Backpropagation~\cite{gbp}, Full-Grad~\cite{fullgrad}, and Grad-CAM~\cite{gradcam}, aim to attribute model predictions to individual spatial feature. More recently, hybrid methods that integrate multiple attribution paradigms have also been proposed~\cite{aggattri}, reflecting efforts to combine the complementary strengths of these techniques.

Despite their intuitive appeal, spatial attribution methods still face challenges in rigorous evaluation. There is no universally accepted metric for assessing correctness~\cite{Quanshi,fullgrad,LPI,infd,withBenchmark1}, and the presence of counterfactual samples~\cite{harvardTrajectory} raises concerns about the reliability and stability of spatial explanations~\cite{unstable,unstablenips,unstableAAAI,unstabletheo22,unstabletheo23,unstablenips2020}. For instance, some methods produce contradictory explanations for the same model and input~\cite{conflict1,conflict2}, and inconsistencies have been observed across input perturbations~\cite{unstable,unstabletheo23}. To address these concerns, a variety of evaluation metrics have been proposed, including FID~\cite{FID}, IR~\cite{IR}, INFD~\cite{infd}, ROAR~\cite{attbenchmark}, DIFFID~\cite{DiffID2}, and the Deletion-Insertion Game~\cite{fullgrad}. Broadly, these metrics fall into two categories. Game theory-based and ablation-based approaches provide a principled framework for evaluating feature importance, but they often suffer from artifacts introduced by deletion operations in the spatial domain~\cite{LPI,fullgrad,craftartifact,TCAV,FromTCAV}. Axiom-based approaches, on the other hand, define criteria based on theoretical properties; however, there is little consensus on which axioms should be considered foundational, resulting in fragmented and sometimes incompatible evaluation schemes.

The limitations of spatial methods motivate a shift to the frequency domain. The Fourier transform decomposes an input into a linear combination of orthogonal frequency components, enabling a mathematically and physically grounded definition of deletion operations---specifically, zeroing out the signal energy, a procedure well-established in classical signal processing~\cite{ampli-filter,Conv}. Consequently, game theory-based evaluation metrics such as the Deletion-Insertion Game can be implemented in the frequency domain without introducing artifacts. This insight motivates our use of Fourier-based attribution as a more principled and robust alternative to spatial methods.
\subsection{Studies of Analyzing the networks via Fourier Features}
An increasing body of work has examined neural networks through the frequency domain, revealing that modern architectures often prioritize low-frequency components~\cite{vitlowfre,wang2022antioversmoothing}. Several studies~\cite{vitgood,vitlowfre,NTKFourier,shapFre} indicate that low-frequency features tend to generalize better, and evidence from compression theory~\cite{compresstheory} and model compression studies~\cite{modelcomression} further supports the notion that low-frequency components carry more information than high-frequency ones. Moreover, convolutional layers have been shown to exhibit implicit frequency-selective behavior~\cite{wen2024which}. However, these theoretical results are often derived under idealized conditions---such as kernels matching the input size or recurrent convolution operations—which are rarely satisfied in contemporary architectures like Vision Transformers (ViTs) and ResNet.

While these studies provide important insights, they generally focus on aggregated frequency properties rather than enabling feature-level attribution, i.e., determining how individual Fourier components contribute to a network's prediction. This represents a significant gap for practical interpretability. In response, we propose the Fourier Feature Correlation (FFC) framework, which extends prior insights into a scalable, architecture-agnostic method capable of attributing the importance of individual Fourier features to model decisions. FFC builds upon classical signal decomposition principles while addressing the limitations of existing spatial attribution methods, offering a principled and robust tool for interpreting modern deep networks.

\section{Method}
\subsection{Preliminary}
Unlike most AI tasks, feature attribution lacks ground-truth labels (because obtaining such labels would require an accurate interpretation of the model, which is precisely the goal of the task itself), compelling researchers to design ablation-based evaluation criteria to assess an attribution method’s faithfulness. The widely accepted definition of a non-important feature underlying these metrics can be expressed as follows:
\begin{equation}
    dis\left(\mathcal{M}\left(\{X\}\right),\mathcal{M}\left(\{X\}/\{X_{non}\}\right)\right) \rightarrow 0,
\end{equation}
where $\mathcal{M}$ denotes the black-box model, $\{X\}$ denotes the sample with all features and $\{X_{non}\}$ denotes the non-important feature. The operator $/$ denotes the deletion operation. However, the situation is more nuanced in practice: The $\{X_{non}\}$ typically exerts a small---rather than strictly zero-impact on the model’s output. 
Thus, current attribution methods tend to assign an attribution score to each feature. The ideal situation of the scores needs to be satisfied:

For a feature score $s_i$ assigned by an attribution method, the method is considered faithful if $s_i \ge s_j$ implies
\begin{equation}
    \mathcal{C}\!\left(\mathcal{M}\!\left(\{X\}/x_i\right),\, \mathcal{M}\left(\{X\}\right)\right)
\;\ge\;
\mathcal{C}\!\left(\mathcal{M}\!\left(\{X\}/x_j\right),\, \mathcal{M}\left(\{X\}\right)\right),
\end{equation}
where $\mathcal{C}$ is the criterion function that quantifies the influence induced by the deletion, $x_i$ denotes one of the features that constitute $\{X\}$.

Because a neural network always handles unstructured data consisting of thousands of features and the features themselves may have interaction effects, the deletion-insertion game~\cite{fullgrad} is proposed to approximately evaluate the performance of an attribution method. 

For a given sorted score sequence of the features, the deletion-insertion game value is computed by:
\begin{equation}
     value = \sum\limits_{s_{min}}^{s_{max}} \mathcal{C}\left(\mathcal{M}, \{X\}/\bigcup_{i=s_{min}}^{i=s_{max}} x_{i}\right).
\end{equation}
The main obstacle in feature attribution is the non-effect operator $/$, which renders ablation studies invalid. However, in the frequency domain, the $/$ operation is well-defined making the evaluation reliable.

\subsection{The Reliability of Fourier Features}

Assume that $X\subset \mathbb{R}^n$ denotes the vector representation of a sample with $n$ features, and the $x_i$ denotes the deleted feature. The deletion operator in the spatial domain can be expressed below:
\begin{equation}
    \{X\}/x_i = X-\left(0,\dots,x_i-v_b,\dots,0\right),
\end{equation}
where $\left(0,\dots,x_i,\dots,0\right)$ is a vector of the same dimension as $X$ whose only non-zero entry is $x_i$ at position $i$, $v_b$ denotes the baseline value. However, ablation studies in numerical AI research typically treat $\{X\}/x_i$ as an element of $\mathbb{R}^{n-1}$ where the feature $x_i$ is completely removed, including its corresponding dimension. These two operators are equivalent only when the non-effect baseline is clearly defined. 

The primary challenge lies in the fact that a well-defined non-effect baseline is exceedingly difficult to establish, which in turn leads to the failure of the deletion operator in the spatial domain. A lot of works criticized this operator for introducing new artifacts~\cite{craftartifact,fullgrad,LPI}. However, this difficulty is effectively overcome in the frequency domain, where the deletion baseline is rigorously defined by setting the spectral power to zero~\cite{ampli-filter}. 

After the Fourier transformation, the sample is represented by a linear combination of a set of mutually orthogonal basis functions: 
\begin{equation}
\label{eq:Xfdel}
    X_f = \sum A_ie^{j\left(\omega_i t+\phi_i\right)},
\end{equation}
where $A_i$ denotes the magnitude, $\omega_i$ denotes the frequency, and $\phi_i$ denotes the phase of the $i$-th signal. 
Thus, setting the corresponding spectral power (magnitude of a Fourier item) to zero aligns with the definition of deletion in signal filtering without influencing other orthogonal components~\cite{ampli-filter}. We represent the deletion operation in the frequency domain as:
\begin{equation}
    \{X_f\}/A_ie^{j\left(\omega_i t+\phi_i\right)} = \{X_f\} - A_ie^{j\left(\omega_i t+\phi_i\right)}.
\end{equation}
From both the signal decomposition and the mathematical perspectives, the Fourier feature $A_ie^{j\left(\omega_i t+\phi_i\right)}$ is eliminated. Mathematically, the number of terms in Eq.~\ref{eq:Xfdel} decreases from $n$ to $n-1$, which is fundamentally different from spatial deletion. From the signal decomposition perspective, the spectral power of the $i$-th component is zeroed out, thereby eliminating its physical influence.

Grounded in the theoretical foundations of signal removal in filtering (\cite{Conv,ampli-filter}), we argue that applying the game theory-based Deletion-Insertion Game in the frequency domain is both mathematically and physically well-justified.
\begin{figure*}
    \centering
    \includegraphics[width=0.9\linewidth]{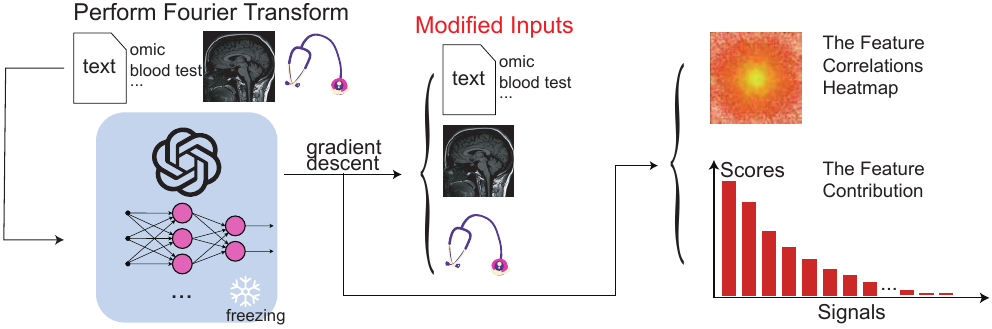}
    \caption{\textbf{The Schematic workflow of FFC}: The original signal is regarded as an adjustable parameter and input to a trained neural network with all parameters frozen. Loss function is leveraged to revise the original signal according to the preferences of the neural network, yielding the Expected Signal. The pair of signals will then undergo a Discrete Fourier Transform and be transformed from the spatial domain to the frequency domain, respectively. Further in the frequency domain, the expected signal will be projected to the original signal, and a measurement of correlation will be conducted. Finally, according to the correlation of each signal pair, we can score and rank the signals in the frequency domain.}
    \label{fig:method}
\end{figure*}
\subsection{Comprehensive Deletion-Insertion Game}
Typically, for classification tasks, we adapt the $\mathcal C$ in deletion-insertion game to compute a subtraction of the relative confidence score after perturbation from $1$, as higher confidence indicates that the corresponding features are more relevant to the target output. The formula can be expressed as:
\begin{equation}
    \mathcal{C}\left(\mathcal{M},X,Q_I\right) = 1-\frac{\mathcal{M}\left(\{X\}/\bigcup\limits_{x_i\in Q_i} x_i\right)}{\mathcal{M}\left(\{X\}\right)}.
\end{equation}
where $Q_I$ is an indication set, and the index $i$ in $Q_I$ indicates that the corresponding feature $x_i$ should be deleted. For convenience, $x_i$ here denotes the features in frequency domain. As deduced above, for Fourier features, the union operator $\bigcup$ is equivalent to the summation $\sum$. Therefore, in the rest of the paper, we use $\sum$ to denote the union operator of Fourier features.
The relative confidence score is defined as the ratio between the original prediction confidence and the confidence obtained after deleting or inserting a subset of features. This score is recalculated at each deletion (or insertion) step corresponding to a specific feature removal rate. We subsequently compute the area under the curve (AUC) of the relative confidence trajectory. 

Inspired by Yang et al.\cite{LPI}, to comprehensively evaluate attribution methods, we quantify performance by subtracting the AUC obtained from deleting the least important features from that obtained by deleting the most important ones. Accordingly, a larger AUC indicates better attribution performance.\\
The evaluation formula can be expressed as:
\begin{align}
\label{eq:AUC}
    AUC= \sum\limits_{s=min\left(\mathcal{A}\left(x\right)\right)}^{s=max\left(\mathcal{A}\left(x\right)\right)} \mathcal{C}\left(\mathcal{M},\{X\}/\sum\limits_{i=min\left(\mathcal{A}\left(x\right)\right)}^{i=s}x_{i}\right)-\nonumber\\\sum\limits_{s=max\left(\mathcal{A}\left(x\right)\right)}^{s=min\left(\mathcal{A}\left(x\right)\right)} \mathcal{C}\left(\mathcal{M},\{X\}/\sum\limits_{i=max\left(\mathcal{A}\left(x\right)\right)}^{i=s}x_{i}\right),
\end{align}
where $s$ denotes the corresponding importance score of the feature $x_i$. \textbf{Equipped with a clear and reliable evaluation metric, we are now able to formally define our problem.}

\subsection{Problem Formation}
Given a sample $X$ consisting of $n$ features, we apply Fourier transform to this sample to obtain $n$ Fourier features $x_i$. For a given black-box model $\mathcal{M}$, we aim to find an ordered sequence of scores $S=\{s_0,\dots,s_n\}$ to maximize the Eq.\ref{eq:AUC}. This procedure can be formulated as:
\begin{equation}
    S^*=\arg \max_S\{AUC\left(\mathcal{M},\{X\},S\right)\}.
\end{equation}
However, solving this equation exactly is an NP-hard problem; therefore, we propose a fast yet accurate algorithm.

\subsection{Fast Fourier Correlation Algorithm}
Directly applying complex-valued optimization is challenging and inefficient for a neural network operating in the real domain. We leverage gradient descent and feedback control to establish a connection between real and complex representations. We model the attribution process as a control process of the closed-loop feedback system. While the idea of modeling networks using control theory dates back to early works on Hopfield networks~\cite{neurondynamics}, to the best of our knowledge, control-theoretic formulations have not yet been applied to feature attribution.

Unlike in network training, where the control parameters are typically the network weights~\cite{analyneurondynamics1}, the attribution process respectively treats the input sample and the model’s output as controllable and observable states. As illustrated in Figure~\ref{fig:method}, we freeze the weights of the target model and treat the attribution process as a dynamic system that iteratively adjusts the input to minimize the discrepancy between the network’s current output and a target output.

As shown in Figure~\ref{fig:method}, we treat the loss function as a controller, with the loss value serving as the control signal. Gradients computed from the network’s output are propagated backward to the input layer, acting as error signals that guide subsequent updates. This mechanism is implemented via gradient descent, enabling real-time estimation of feature importance in the frequency domain.
Then we adopt the gradient descent algorithm and error signals to rectify the input signals and eliminate the loss value. We refer to the resulting modified inputs—which steer the network output toward a steady state that minimizes the loss---as the ``expected signal”. From a control-theoretic perspective, we assume that the original signals consist of both expected signals and noise. During the modification process, expected signals are enhanced, while noise signals are suppressed or substantially altered. Consequently, the correlation between the original signals and the expected signals serves as a measure of signal importance. To evaluate the contribution of each signal independently, we transform both sets of signals into the frequency domain (K-space) and compute the projection of each component of the expected signals onto the corresponding component of the original signals. The projection formula is given below:
\begin{align}
    F_{X'} &=\mathcal{F}\left(X^{\left(n\right)}-lr\cdot\nabla X^{\left(n\right)}\right)\\
    Proj\left(u,v\right) &= \frac{2\times Re\left(F_X\left(u,v\right)\cdot\overline{F_{X'}}\left(u,v\right)\right)}{Mag\left(F_{X}\left(u,v\right)\right)},
\end{align} where $lr$ denotes the learning rate, $X^{\left(n\right)}$ denotes the rectified input after $n$ iterations of gradient descent algorithm, $X^{\left(n\right)} - \nabla X^{\left(n\right)}$ denotes the expected signals $X'$, $\mathcal{F}$ means Fourier transformation, $\overline{F_{X'}}$ means the conjugation signal of the expected signal $F_{X'}$ and the $Re\left(\cdot\right)$ means the real part of a complex number, $Mag\left(\cdot\right)$ denotes the magnitude of the signals. And then we use the value of the projection minus the magnitude of the original signal to obtain the importance score:
\begin{equation}
    \mathcal A\left(u,v\right) = Proj\left(u,v\right)-Mag\left(F_X\left(u,v\right)\right),
\end{equation}
The projection captures the correlation between the expected signals and the original signals. To mitigate the effect of scale discrepancies in signal magnitude, we subtract the original magnitude. The overall procedure is presented in the Algorithm \ref{alg:ffc}, where $X$ denotes the original input, $\mathcal M$ denotes the neural network, $lr$ denotes the learning rate, and $E$ denotes the number of iterations.

\begin{algorithm}[H]
\caption{Fast Fourier Correlation.}\label{alg:ffc}
\begin{algorithmic}
\STATE {\textbf{Inputs:} $X$, $\mathcal M$, $lr$, $E$}
\STATE {\textbf{Outputs:} $\mathcal{A}\left(X\right)$}
\STATE
\STATE {\textsc{Obtain Expected Signals}} $(X')$
\STATE \hspace{0.5cm}\textbf{For} $e$ \textbf{from} $0$ \textbf{to} $E$ \textbf{Do}:  
\STATE \hspace{1.0cm} $X^{\left(e+1\right)} = X^{\left(e\right)}-lr\cdot\frac{\partial M}{\partial X^{\left(e\right)}}$
\STATE 
\STATE {\textsc{Get Correlation}} $(\mathcal{A}\left(X\right))$
\STATE \hspace{0.5cm}$F_{X'}=FFT\left(X'\right)$
\STATE \hspace{0.5cm}$F_{X} = FFT\left(X\right)$
\STATE \hspace{0.5cm}$Proj\left(u,v\right) = \frac{2\times Re\left(F_X\left(u,v\right)\cdot\overline{F_{X'}}\left(u,v\right)\right)}{Mag\left(F_{X}\left(u,v\right)\right)}$
\STATE \hspace{0.5cm}$\mathcal A\left(u,v\right)=Proj\left(u,v\right)-Mag\left(F_X\left(u,v\right)\right)$
\STATE \textbf{return} $\mathcal A\left(X\right)$
\end{algorithmic}
\label{alg1}
\end{algorithm}
According to the definition and derivation of the Input-Grad-like attribution methods~\cite{intgrad}, FFC satisfies \emph{Implementation Invariance}. Assume that two black-box neural networks $f_1\left(\cdot\right),f_2\left(\cdot\right)$ have equivalent outputs but have different implementations. Having equivalent outputs means that for any input $x$, $f_1\left(x\right)=f_2\left(x\right)$. According to the definition of the gradient, we have:
\begin{align}
    \frac{\partial f_1}{\partial x} &= \lim_{\Delta x\rightarrow 0}\frac{f_1\left(x+\Delta x\right)-f_1\left(x\right)}{\Delta x} \nonumber\\&= \lim_{\Delta x\rightarrow 0}\frac{f_2\left(x+\Delta x\right)-f_2\left(x\right)}{\Delta x} \nonumber\\&= \frac{\partial f_2}{\partial x}. 
\end{align}
Thus, for any two equivalent black-box networks, their gradients are equal, and FFC consequently satisfies \emph{Implementation Invariance}.

\section{Experiment}
Since faithfulness and sensitivity are two important criteria of attribution methods, the experiments are designed to address the following research questions:
\begin{itemize}
    \item How do learning rate and number of iterations influence the results?
    \item Does FFC exhibit greater faithfulness compared to traditional spatial-domain attribution methods?
    \item Are all the steps of FFC essential?
    \item Is FFC sensitive to noise?
    \item How does the computational overhead of FFC compare with existing attribution approaches?
    \item Does FFC support multiple modalities and models?
    \item What properties do Fourier features exhibit, particularly for the high-scoring ones?
\end{itemize}
We arrange our experiments as follows:\\
\textbf{Find suitable parameters:}We begin by analyzing the impact of learning rate and the number of iterations on FFC, in order to determine the optimal parameters for the subsequent experiments. \\
\textbf{Superiority over conventional algorithms:} We compare the faithfulness of FFC with that of baseline methods under the DIG metric to demonstrate the superiority of FFC. The ablation study demonstrates that every step of our design is crucial and our superiority can not be obtained by any trivial transformation of existing methods. In addition, we conduct a sensitivity analysis following prior work~\cite{infd,aggattri}, which reveals that FFC maintains robust performance under noisy conditions. Extensive experiments on generated datasets, along with adversarial experiment, further demonstrate the faithfulness of FFC.\\
\textbf{The generability of FFC:} We first demonstrate that FFC is computationally demanding in terms of both time and memory when applied to large-scale datasets. We then conduct the experiments on different modalities and models showing that FFC can be applied to modern tasks and structures.\\
\textbf{The Characteristics of Fourier Features:} We analyze the \textbf{\emph{structural characteristics}} of high-scoring Fourier features, showing that they exhibit more fixed order and positional stability while maintaining comparable specificity. Furthermore, we conduct error rectification experiments to visualize the detail information carried by attributed features, illustrating that Fourier features encode precise semantic content beyond the reach of saliency maps. 

\subsection{Settings}
\textbf{Dataset}: Validation set of the ImageNet2012~\cite{imgnet2012} dataset with $1000$ categories ($50$ samples per category) and CIFAR-10. These datasets have been used in many previous works~\cite{fullgrad,craftartifact,LPI,gradcam}. The complete Imgnet2012 validation set and CIFAR-10 are interpreted in terms of individual baselines.\\
\textbf{Baselines}: Integrated-based: Int-Grad~\cite{intgrad}, Perturbation-based: Smooth-Grad~\cite{smoothgrad}, Back-Propagation: Grad-CAM~\cite{gradcam}, Full-Grad~\cite{fullgrad}, Input-Grad~\cite{inputgrad}. We select baselines from different types of foundations.\\
\textbf{Implementation details}: Experiments are conducted with A800 80GB and CentOS 8. The IG is implemented by captum ~\cite{captum}. The baseline value is set to zero. Other baselines are implemented by open-source code of~\cite{fullgrad}. All parameters are set according to the default parameters from open-source code and the relevant research paper. The discrete Fast Fourier transforms (FFT) are implemented by PyTorch's native torch.fft package. All experiments requiring random variables were repeated three times, and the average result was taken. \textbf{NOTE: Except for Smooth-Grad, the other methods DO NOT contain random variables, thus the errors are too small. we use the best result of Smooth-Grad for the experiments}. For details on the error, refer to our supplementary. As for Full-Grad in ImageNet2012, because we didn't retrain the ResNet50, the accuracy of the official ResNet-50 is around $80\%$, the result is different from their original paper (their accuracy is above $95\%$). \textbf{Full-Grad does not support ViT\_B/32.}\\
\textbf{The backbone networks}: ResNet-50 and ViT-B/32 for ImageNet2012, both use the official PyTorch pre-trained models with default initialization weights. ResNet-18 and VGG16 for CIFAR-10.
\subsection{Parameter Analysis}
\begin{figure*}
    \centering
    \subfloat[]{\includegraphics[width=0.25\linewidth]{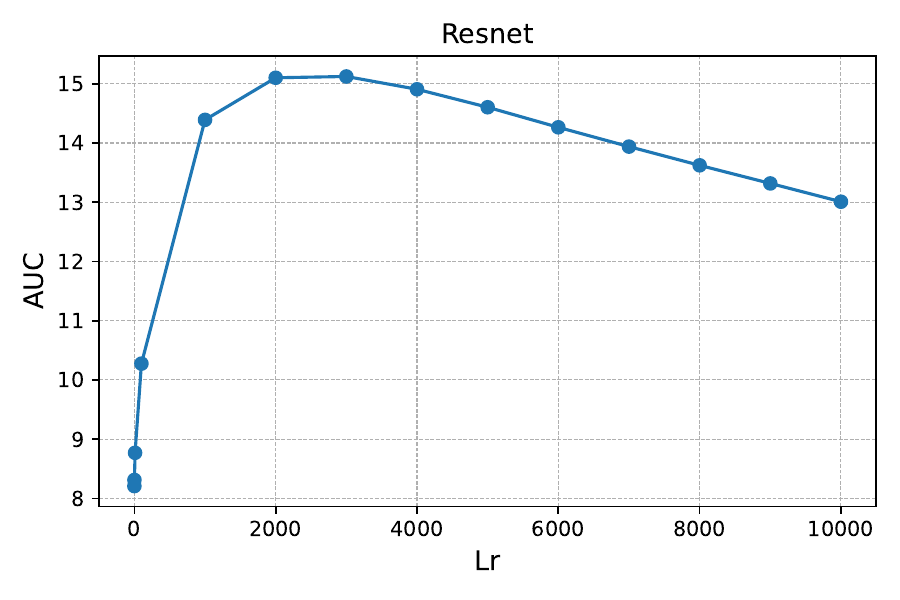}%
    \label{fig:Lr-Resnet}}
    \hfil
    \subfloat[]{\includegraphics[width=0.25\linewidth]{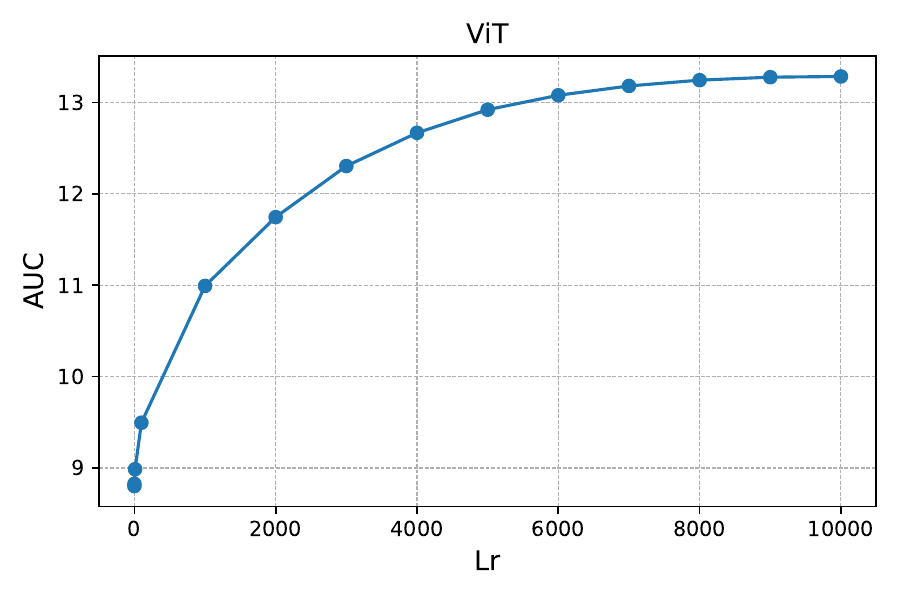}%
    \label{fig:Lr-ViT}}
    \hfil
    \subfloat[]{\includegraphics[width=0.25\linewidth]{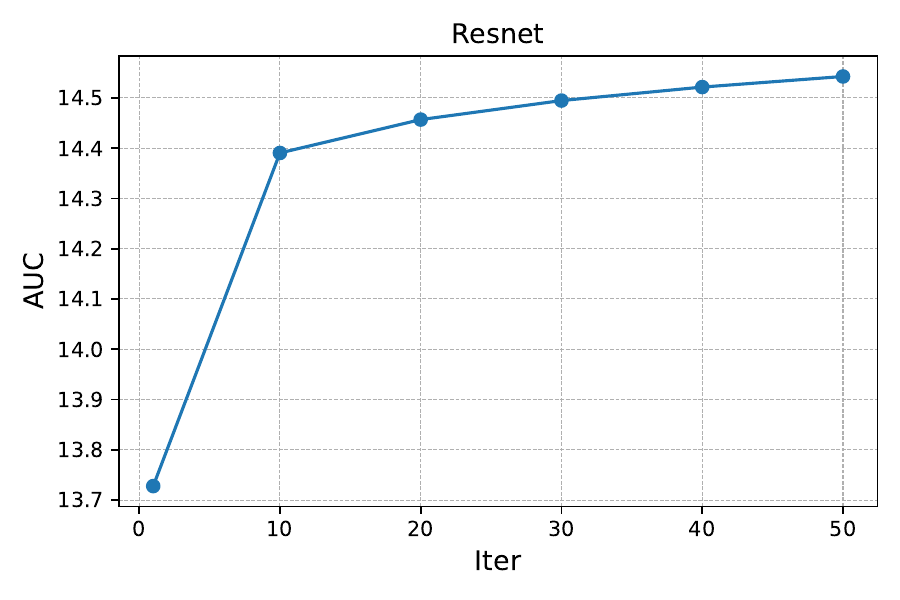}%
    \label{fig:Iter-Resnet}}
    \hfil
    \subfloat[]{\includegraphics[width=0.25\linewidth]{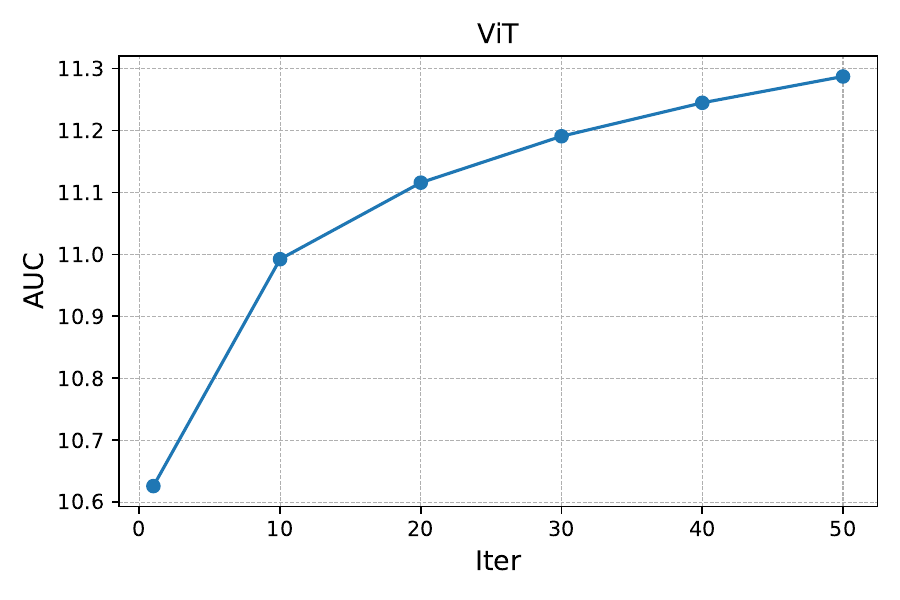}%
    \label{fig:Iter-ViT}}
    \caption{Parameter Analysis: For the learning rate analysis, the number of iterations is fixed at $5$. For the iteration analysis, the learning rate is fixed at $1000$. The y-axis denotes the AUC value of the DIG.}
    \label{fig:param-ana}
\end{figure*}
\label{Param-Ana}

FFC is influenced by both the learning rate and the number of iterations. This section analyzes their impact on the AUC score of the Deletion–Insertion Game. With the gradient norm ignored~\cite{GD}, we vary the learning rate ($lr$) from $0.1$ to $10,000$ and the number of iterations from $1$ to $50$. As shown in Figure~\ref{fig:Lr-Resnet} and Figure~\ref{fig:Lr-ViT}, rather than introducing instability, the AUC increases rapidly as the learning rate increases. Moreover, with an increasing number of iterations, as shown in Figure~\ref{fig:Iter-Resnet} and Figure~\ref{fig:Iter-ViT}, the AUC of FFC exhibits a stable decreasing trend even when $lr > 1$. When the iteration count exceeds $1$, the AUC increases slightly, but the effect is not statistically significant. These results indicate that FFC is primarily influenced by the learning rate rather than the number of iterations. \textbf{FFC performs consistently well across a wide range of parameters. It shows low sensitivity to the parameters.} Thus, we adopt $lr=1000, e=5$ to conduct the rest of the experiments.

\subsection{FFC Exhibits Greater Faithfulness}
\textbf{High-resolution Dataset ImgNet-2012, Modern and Traditional Net Structures:}
To compare the differences between FFC and conventional spatial attribution algorithms under the Deletion---Insertion Game (DIG) metric in their originally designed domain, we sequentially set pixel values in the spatial domain to zero---a widely adopted removal operation in prior works~\cite{fullgrad,DiffID2,LPI,EG,FID,attbenchmark}---according to the importance scores assigned by spatial attribution methods. To ensure consistency across the two domains, we equalize the number of parameters removed (pixels in the spatial domain versus signals in the frequency domain). Accordingly, for FFC, we remove an equivalent amount of signals in the frequency domain based on the attribution scores.

\begin{figure*}
    \centering
    \subfloat[]{\includegraphics[width=0.48\linewidth]{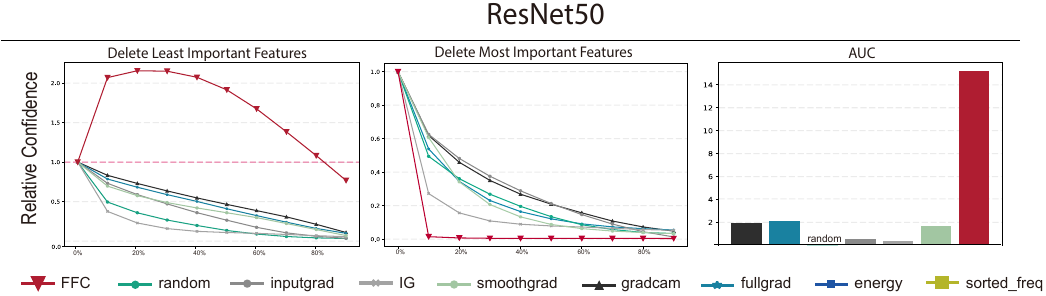}%
    \label{fig:DIG-Resnet-Spatial}}
    \hfil
    \subfloat[]{\includegraphics[width=0.48\linewidth]{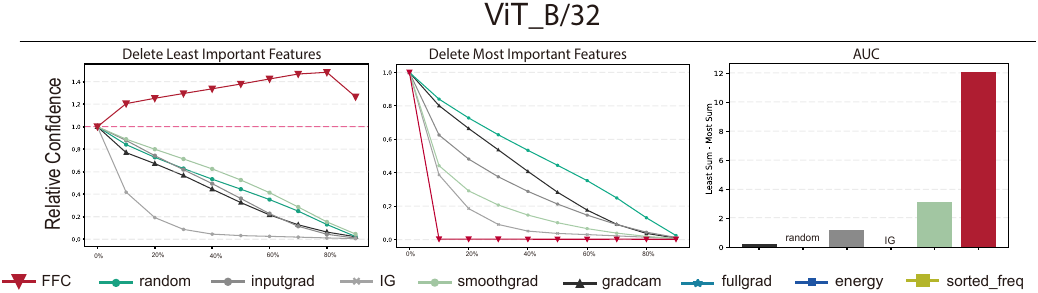}%
    \label{fig:DIG-ViT-Spatial}}
    \hfil
    \subfloat[]{\includegraphics[width=0.48\linewidth]{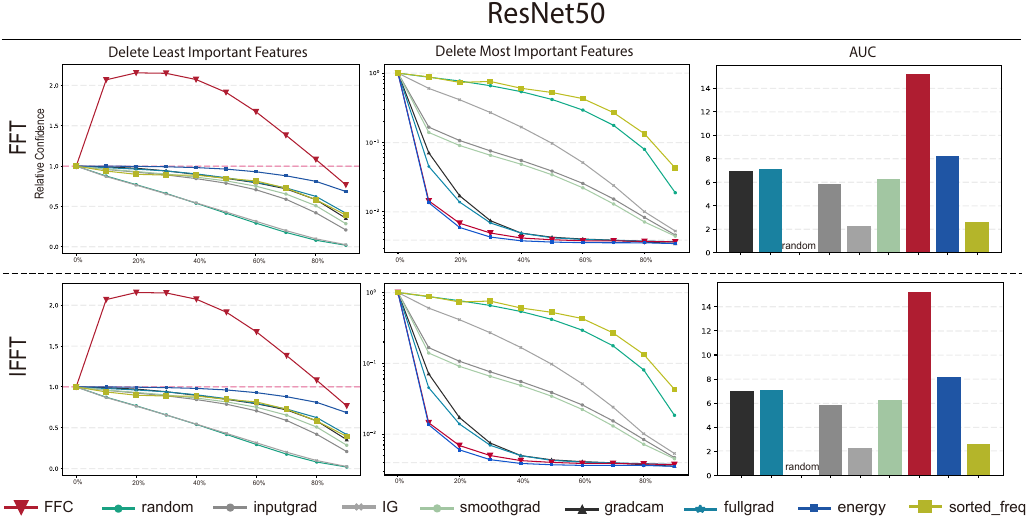}%
    \label{fig:DIG-Resnet-FFT}}
    \hfil
    \subfloat[]{\includegraphics[width=0.48\linewidth]{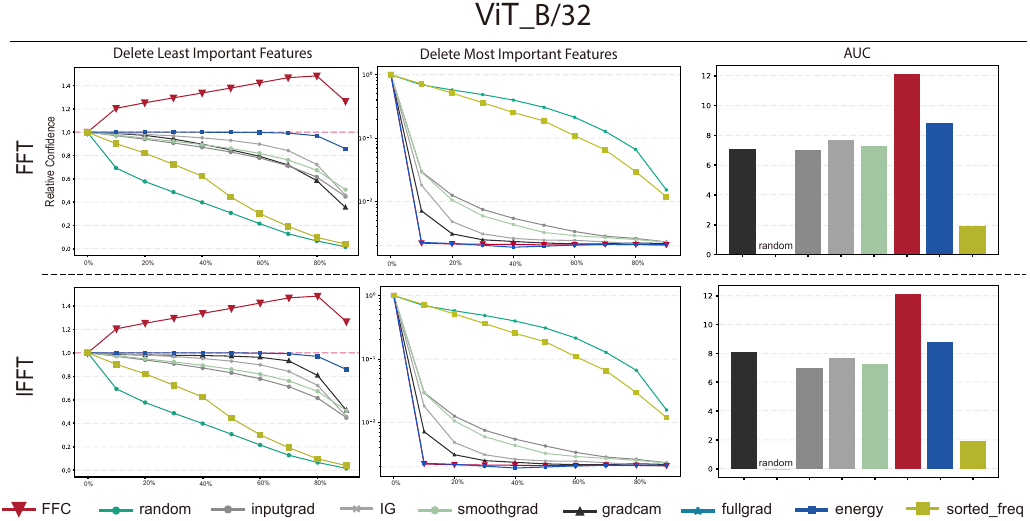}%
    \label{fig:DIG-ViT-FFT}}
    \caption{DIG Results on ImageNet-2012. The curves illustrate the performance trends of different attribution methods as the most or least important features are removed, while the bars summarize the corresponding AUC values of these methods. Curves: the x-axis represents the percent of features being filtered out, the y-axis represents the confidence of the network. Bar plots: the x-axis indicates attribution methods, the y-axis represents the AUC.}
    \label{fig:DIG}
\end{figure*}

As shown in the Figure~\ref{fig:DIG-Resnet-Spatial} and Figure~\ref{fig:DIG-ViT-Spatial}, removing the low-score features identified by FFC can increase the network's confidence. This indicates that the low-score features identified by FFC are noise that confuses the network. In contrast, none of the spatial attribution algorithms can achieve an increase in network confidence. Moreover, even though FFC removes up to $90\%$ of the low-score signals, the network's relative confidence is still around $1$. In experiments removing the most important features, FFC’s relative confidence decreased sharply, approaching zero with the deletion of just $10\%$ of the features. This demonstrates that FFC can eliminate most of the noise while retaining the important information. This fully illustrates that FFC significantly outperforms spatial attribution algorithms in feature selection capability. Additionally, the AUC value significantly surpasses the baselines, demonstrating that FFC is more faithful. It is worth noting that the AUC of random removal is close to zero; however, in the separate deletion game, random removal may surpass certain baselines. This observation suggests that subtracting the deletion AUC from the insertion AUC serves as an effective adjustment to mitigate the bias, making the metric more stable.\\
\textbf{Low-Resolution Dataset CIFAR-10, Traditional Net Structures:}
As illustrated in Table~\ref{tab:DIG-CIFAR-10}, the AUC value of FFC outperforms all other baselines, indicating that FFC can recognize the most influential Fourier features, not only on high-resolution datasets but also on low-resolution ones.

The results of the different backbone networks and different types of dataset show that, \textbf{1) FFC can accurately identify the most important features that influence network decisions, thereby demonstrating its excellent faithfulness. 2) FFC is compatible with both traditional and modern network architectures, as well as image datasets of varying resolutions. These two experiments effectively exhibit FFC’s property of implementation invariance. }
\begin{table*}[]
    \centering
    \caption{DIG Result for CIFAR-10}
    \label{tab:DIG-CIFAR-10}
    \begin{tabular}{l|cccccc}
\toprule
& \textbf{FFC} & \textbf{Sorted Freq} & \textbf{Energy}
& \textbf{Grad-CAM(Spatial)} & \textbf{Input-Grad(Spatial)} & \textbf{Full-Grad(Spatial)}
\\ \midrule
\textbf{ResNet} & \textbf{6.59} & 2.79 & 6.34 & 2.13 & 0.19 & 2.87 \\
\textbf{VGG}    & \textbf{6.76} & 2.37 & 6.26 & 0.04 & 0.07 & 3.00 \\
\midrule
& \textbf{Smooth-Grad(Spatial)} & \textbf{IG(Spatial)}
& \textbf{Grad-CAM(FFT)} & \textbf{Input-Grad(FFT)} & \textbf{Full-Grad(FFT)} & \textbf{Smooth-Grad(FFT)}
\\ \midrule
\textbf{ResNet} & 0.84 & 2.62 & 3.23 & 2.34 & 3.19 & 1.36 \\
\textbf{VGG}    & 0.77 & 1.95 & 3.35 & 2.13 & 3.13 & 1.36 \\
\midrule
& \textbf{IG(FFT)} & \textbf{Grad-CAM(IFFT)} & \textbf{Input-Grad(IFFT)} & \textbf{Full-Grad(IFFT)} & \textbf{Smooth-Grad(IFFT)} & \textbf{IG(IFFT)}
\\ \midrule
\textbf{ResNet} & 0.71 & 3.23 & 2.34 & 3.19 & 1.36 & 0.71 \\
\textbf{VGG}    & 0.46 & 3.35 & 2.13 & 3.13 & 1.36 & 0.46 \\
\bottomrule
\end{tabular}
\end{table*}

\subsection{Ablation Study}

This section underscores the critical role of the correlation step in FFC, emphasizing that its attribution results cannot be trivially obtained from existing methods. To this end, we transform the attribution outputs of baseline methods to the frequency domain using the FFT and IFFT(Inverse Fast Fourier Transform) of their scores, adopting the magnitude of their Fourier-domain representations, and then evaluate them using the Deletion–Insertion Game. Considering the observation of compression techniques \cite{compresstheory,modelcomression, vitlowfre,Fprinciple} and the adversary training techniques \cite{shapFre}, we introduce two additional baselines, which include algorithms that delete features based on frequency magnitude, random scores, or signal magnitude.

As shown in the second and third rows of Figure~\ref{fig:DIG-ViT-FFT}, Figure~\ref{fig:DIG-Resnet-FFT}, and Table~\ref{tab:DIG-CIFAR-10}, FFC consistently outperforms all baseline approaches. More specifically, comparisons with sorted\_freq test FFC against compression techniques, demonstrating superior performance. Similarly, FFC surpasses the energy-based baseline, indicating that its results cannot be explained by previous claims of energy bias driving network preferences. Finally, FFC outperforms the FFT/IFFT transformations of current spatial-domain methods, confirming that its attribution results are not a trivial extension of existing methods. \textbf{Collectively, these results underscore the critical role of the correlation step in FFC, as well as the novelty of FFC over prior attribution approaches.}

\subsection{Faithfulness Experiments on Generated Dataset}
Although the ground truth cannot be labeled in practical datasets, we manually generate a dataset with known ground-truth features. Each class is assigned a distinct featured signal, and Gaussian noise with zero mean is added to the samples. We train an MLP on this manually constructed dataset and interpret the model using FFC to determine whether it can correctly identify the featured signals. The variance of the added noise is gradually increased to evaluate FFC’s robustness. For FFC, the features with the top-1 scores are selected as the predicted features. We define interpret accuracy as the proportion of cases where the attributed features exactly match the ground truth; inclusion rate as the fraction of cases where the ground-truth label is among the attributed features; and redundant rate as the ratio of attributed features to ground-truth features.

\begin{figure}
    \centering
    \includegraphics[width=0.8\linewidth]{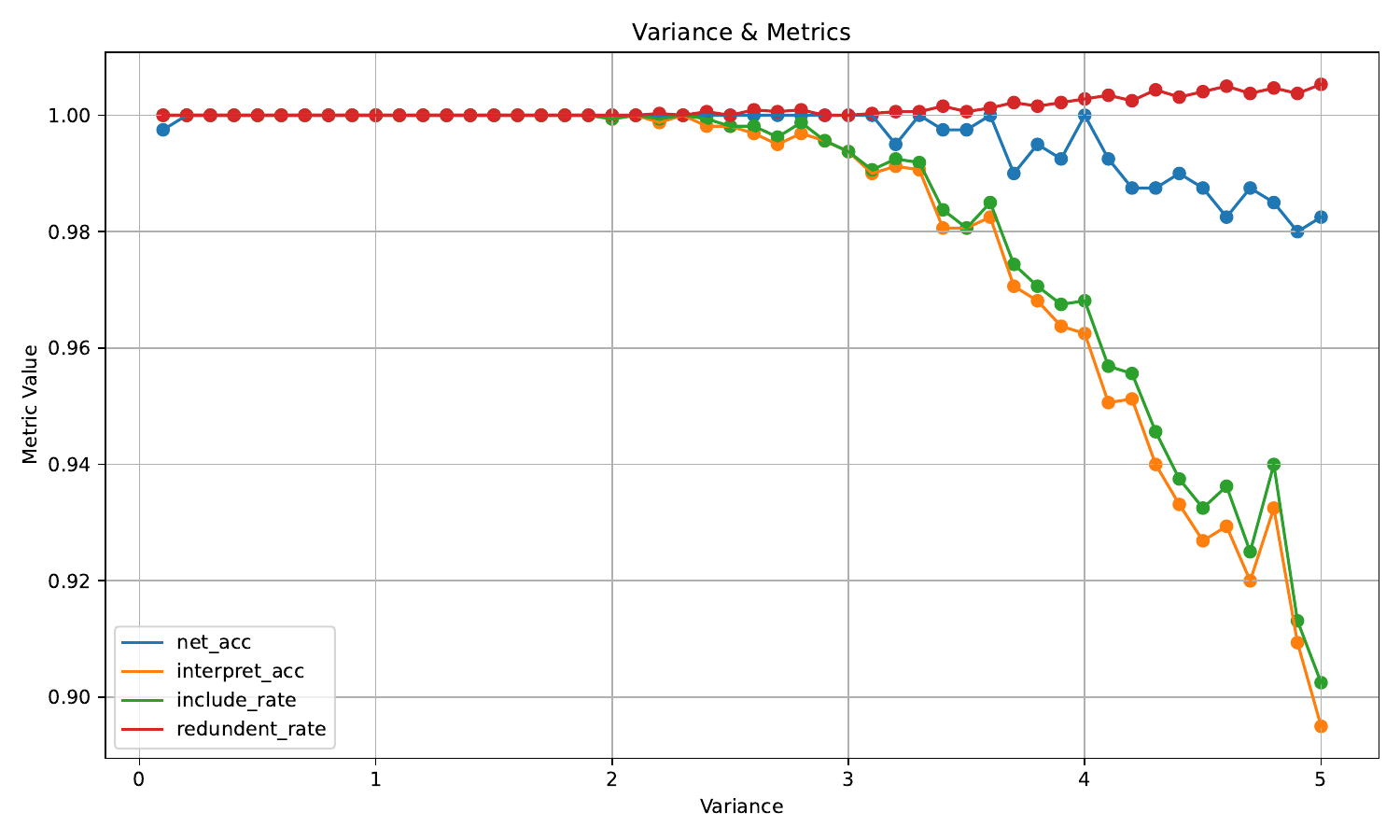}
    \caption{Experiments on a manually constructed dataset with ground-truth label.}
    \label{fig:gt-exp}
\end{figure}

As shown in Figure~\ref{fig:gt-exp}, as the noise variance increases, the network accuracy, interpret accuracy, and inclusion rate gradually decrease, with interpret accuracy and inclusion rate declining faster than the network accuracy. Nevertheless, the interpret accuracy remains above 90\% even when the variance reaches 5, while the magnitude of the information values ranges from -1.8 to 1.8, occasionally reaching up to three times the magnitude of the original information values. Notably, we found that when the noise magnitude exceeds nearly twice the magnitude of the information values, the performance of FFC decreases. These results demonstrate the strong ability of FFC to reliably identify relevant features. 

\subsection{Adversarial Experiments Further Demonstrate the Faithfulness of FFC}
\begin{figure}
    \centering
    \includegraphics[width=0.8\linewidth]{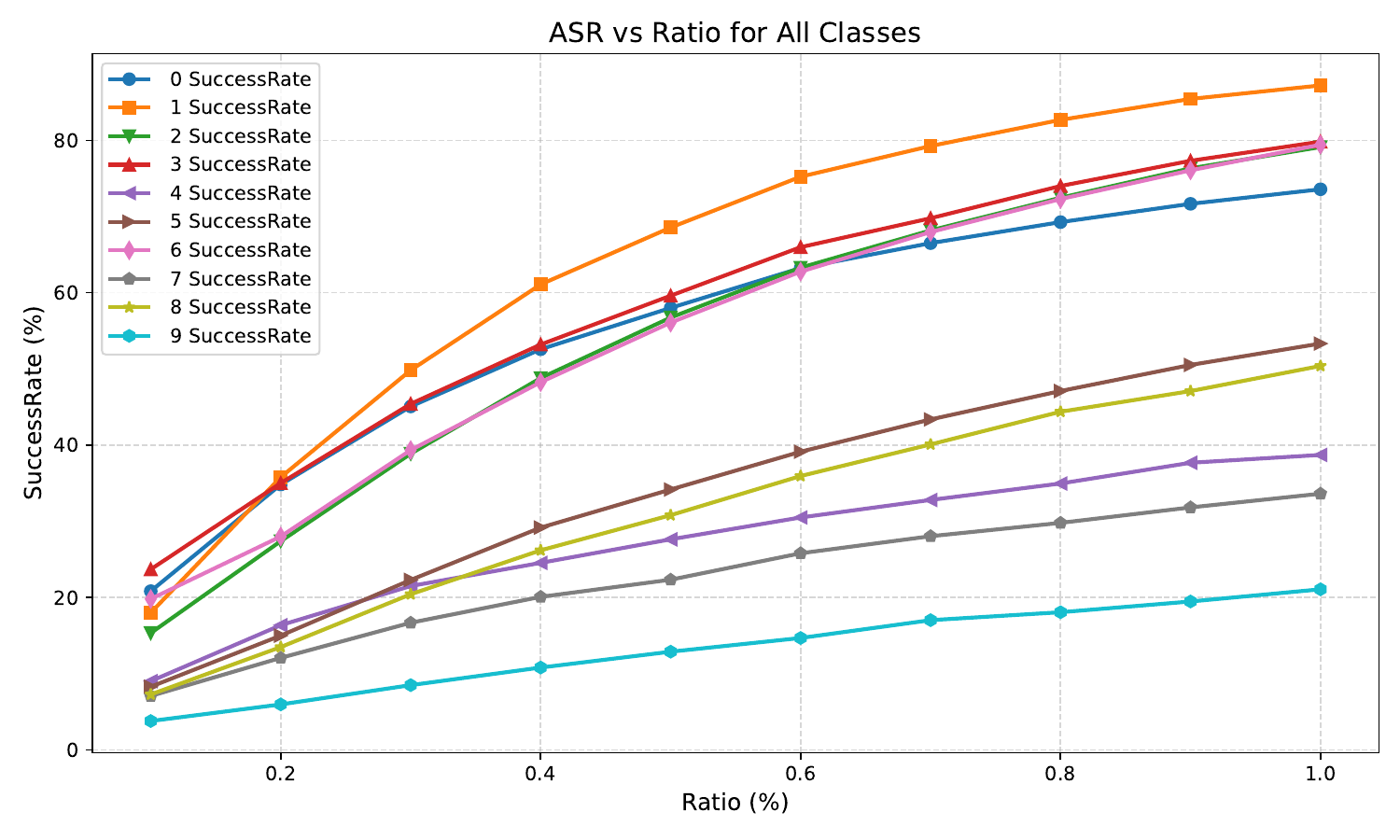}
    \caption{The x-axis denotes the swap rate of the features and the y-axis represents the success rate of shifting network's prediction. An interesting phenomenon is that ResNet-18 exhibits a clear preference among the ten classes, as evidenced by a distinct separation gap across the ten curves.}
    \label{fig:exp-swap}
\end{figure}
We conduct our experiments using a ResNet-18 model trained on CIFAR-10, extracting features from the test set. For each target class, we first collect the features of all samples to build a feature dictionary. Given a sample from another class, we query features according to a specified swap rate. If swapping certain features shifts the model’s prediction to the target class, we consider it a successful shift and compute the overall success rate across the dataset. As shown in Figure~\ref{fig:exp-swap}, for 4 classes, swapping as little as $\mathbf{1\%}$ of the features achieves a success rate approaching 80\%. Only class 9 has a relatively weak success rate, but it can still shift one-fifth of the samples in the dataset with only $1\%$ of the features swapped. These results clearly demonstrate the strong faithfulness of FFC.

\subsection{FFC Exhibits Low Sensitivity to Noise}
To evaluate sensitivity, we applied Gaussian noise to the input layer and computed sensitivity following the approach proposed by Yeh et al.~\cite{infd}, which has been widely adopted for assessing the sensitivity of attribution methods. 

As illustrated in Table~\ref{tab:sens}, on the high-resolution dataset, FFC exhibits extremely low sensitivity (only second to the IFFT of Smooth-Grad), ranging from $0.08$ to $0.13$. On the low-resolution CIFAR-10 dataset, FFC’s sensitivity increases slightly but still ranks third, following only Grad-CAM and Full-Grad. This indicates that FFC is generally robust to input noise. \textbf{Moreover, frequency-domain attribution methods on high-resolution images are better able to resist the impact of noise.}
\begin{table}[ht]
\centering
\caption{Sensitivity scores for CIFAR and ImageNet. The Lower the Better.}
\label{tab:sens}
\begin{tabular}{@{}lS[table-format=2.6]@{\hspace{2em}}lS[table-format=2.6]@{}}
\toprule
\multicolumn{2}{@{}c}{\textbf{VGG16}} & \multicolumn{2}{c}{\textbf{Resnet18}}\\
\midrule
\multicolumn{4}{c}{\textit{CIFAR-10}}\\
FFC               & \textbf{0.106113} & FFC               & \textbf{0.533927}\\
IG Spatial        & 1.187430 & IG Spatial        & 1.465406\\
IG FFT            & 0.756517 & IG FFT            & 1.029796\\
IG IFFT           & 0.756517 & IG IFFT           & 1.029796\\
Grad-CAM Spatial  & 0.171151 & Grad-CAM Spatial  & 0.295480\\
Grad-CAM FFT      & 0.170896 & Grad-CAM FFT      & 0.284764\\
Grad-CAM IFFT     & 0.166733 & Grad-CAM IFFT     & 0.275977\\
Input-Grad Spatial& 1.671534 & Input-Grad Spatial& 1.810410\\
Input-Grad FFT    & 1.100450 & Input-Grad FFT    & 1.174991\\
Input-Grad IFFT   & 1.009770 & Input-Grad IFFT   & 1.120156\\
Smooth-Grad Spatial&0.728722& Smooth-Grad Spatial& 0.771507\\
Smooth-Grad FFT   & 0.406809 & Smooth-Grad FFT   & 0.420426\\
Smooth-Grad IFFT  & 0.260787 & Smooth-Grad IFFT  & 0.275890\\
Full-Grad Spatial & \textbf{0.075680} & Full-Grad Spatial & 0.127985\\
Full-Grad FFT & 0.406809 & Full-Grad FFT & 0.117356\\
Full-Grad IFFT & 0.260787 & Full-Grad IFFT & \textbf{0.117349}\\
\midrule
\multicolumn{2}{@{}c}{\textbf{ViT\_B/32}} & \multicolumn{2}{c}{\textbf{Resnet50}}\\
\midrule
\multicolumn{4}{c}{\textit{ImgNet-2012}}\\
FFC               & \textbf{0.085210} & FFC               & \textbf{0.122036}\\
IG Spatial        & 0.907778 & IG Spatial        & 1.232572\\
IG FFT            & 0.608310 & IG FFT            & 0.666432\\
IG IFFT           & 0.608310 & IG IFFT           & 0.666432\\
Grad-CAM Spatial  & 1.308053 & Grad-CAM Spatial  & 0.825840\\
Grad-CAM FFT      & 61.329509 & Grad-CAM FFT      & 8.822798\\
Grad-CAM IFFT     & 0.254351 & Grad-CAM IFFT     & 0.332242\\
Input-Grad Spatial& 1.259699 & Input-Grad Spatial& 1.250694\\
Input-Grad FFT    & 0.842449 & Input-Grad FFT    & 0.691261\\
Input-Grad IFFT   & 0.322307 & Input-Grad IFFT   & 0.479228\\
Full-Grad Spatial & Not Support & Full-Grad Spatial & 0.257691\\
Full-Grad FFT     & Not Support & Full-Grad FFT     & 0.186974\\
Full-Grad IFFT    & Not Support & Full-Grad IFFT    & 0.186962\\
Smooth-Grad Spatial&0.457038& Smooth-Grad Spatial&0.723322\\
Smooth-Grad FFT   & 0.316964 & Smooth-Grad FFT   & 0.415121\\
Smooth-Grad IFFT  & \textbf{0.054898} & Smooth-Grad IFFT  & \textbf{0.070578}\\
\bottomrule
\end{tabular}
\end{table}

\subsection{FFC Exhibits Low Computational Overhead}
To compare the computational overhead of FFC with existing attribution algorithms, we fix the batch size to $128$ to compare the spatial and temporal overheads. To better highlight the differences among methods and to more closely reflect practical applications, we evaluate the computational overhead on high-resolution datasets. For methods that exceed the device limits, we use the maximum batch size that we can handle. The spatial overhead of the algorithms is measured by the peak memory usage during runtime, while the temporal overhead is measured by the average time required to compute one batch. The representation method is average time × iteration ($e$) times. To eliminate errors caused by system I/O, the time cost is measured from the moment the data enters the GPU. Given that FFC is mainly affected by the learning rate, we set $e=1$ in this experiment. Despite this, Figure~\ref{fig:param-ana} shows that FFC continues to outperform its spatial-domain counterparts.
\begin{table*}[]
    \centering
    \caption{Computational Overhead}
    \label{tab:cost}
    \begin{tabular}{c|c|c|c|c}
    \toprule
       Methods  & Batch Size & Time(s/Batch$\times$Batch Num) & Space(MB) & Model \\
       \hline
       IG  & 16 & 1.422$\times$3125  & 73010 & \multirow{8}*{ResNet50} \\
       Full-Grad  & 32 & 0.124s$\times$1563  & 51278 & ~ \\
       Smooth-Grad  & 128 & 8.521s$\times$391  & 12212 & ~ \\
       Input-Grad  & 128 & \textbf{0.067}$\times$391  & 12004 & ~ \\
       Grad-CAM  & 128 & 0.087$\times$391  & 13122 & ~ \\
       FFC (e$=1$,AUC$=15.10$)  & 128 & \textbf{0.071}$\times$391  & 14064 & ~ \\
       \hline
       IG  & 32 & 2.093$\times$1563  & 53206 & \multirow{7}*{ViT\_B\_32} \\
       Smooth-Grad  & 128 & 16.135$\times$391  & 5422 & ~ \\
       Input-Grad  & 128 & \textbf{0.067}$\times$391  & 4908 & ~ \\
       Grad-CAM  & 128 & 0.087$\times$391  & 5684 & ~ \\
       FFC (e$=1$,AUC$=11.75$)  & 128 & \textbf{0.081}$\times$391  & 6490 & ~\\
    \end{tabular}
    
\end{table*}
Compared to the baselines, the time cost of FFC is second only to input*grad when the number of iterations ($e$) is $1$. As displayed in Figure~\ref{fig:param-ana} and Table~\ref{tab:cost}, the AUC value is similar to the $e=50$ (change rate less than $4\%$). The memory cost of FFC is also close to the baselines, except for IG and Full-Grad. Notably, since Input-Grad involves only back-propagation and multiplication once, no method can be faster than it due to its simplicity. 

\textbf{Since the computational overhead of back-propagation is close to that required for model training, we argue that FFC has potential for industry applications.}

\subsection{Fourier Feature Exhibits Structural Characteristics}
To study the characteristics of Fourier features found by FFC, this section analyzes the FFC attribution results by comparing them to existing spatial results from the perspective of intra-class concentration and inter-class specificity. The high-score features of different algorithms are defined as features with scores exceeding the mean score within each sample. Further, to eliminate the differences of scale across methods, we tag the high-score feature to $1$ and the low-score to $0$. 
\begin{figure*}
    \centering
    \subfloat[]{
    \includegraphics[width=0.23\linewidth]{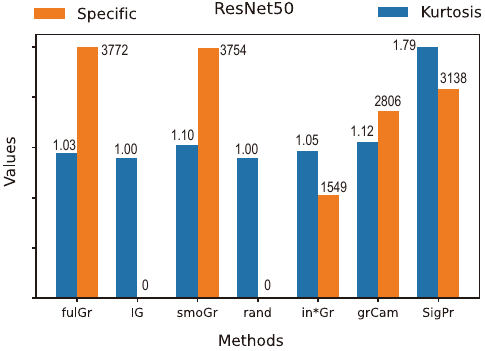}
    \label{fig:structureRes}
    }
    \hfil
    \subfloat[]{
    \includegraphics[width=0.23\linewidth]{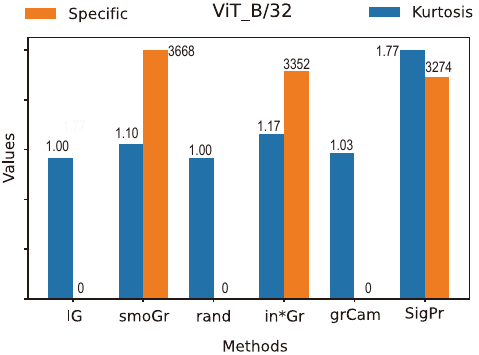}
    \label{fig:structureViT}
    }
    \hfil
    \subfloat[]{
    \includegraphics[width=0.23\linewidth]{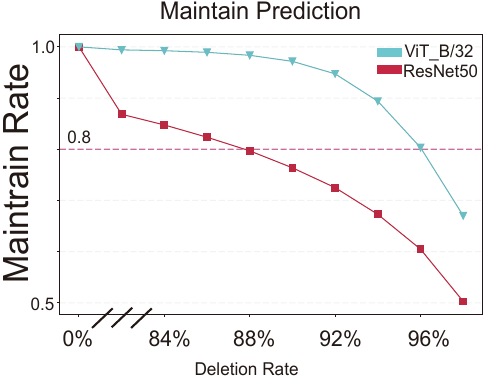}
    \label{fig:minimum}
    }
    \hfil
    \subfloat[]{
    \includegraphics[width=0.23\linewidth]{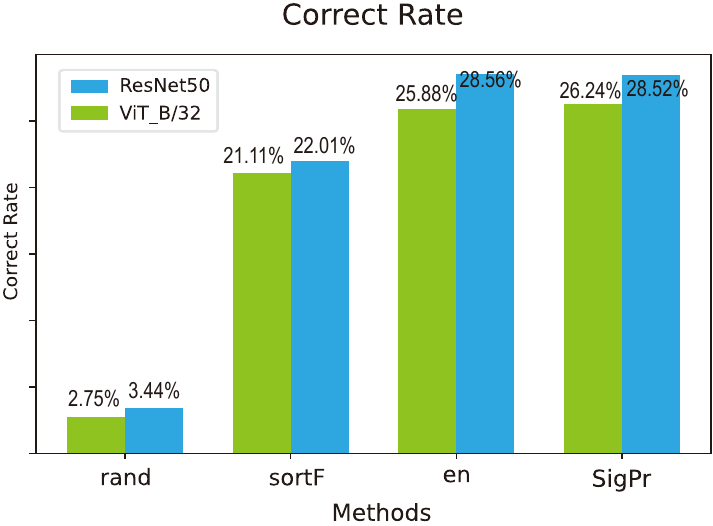}
    \label{fig:cor-rate}
    }
    \caption{(a-b) Intra-class Concentration (blue) and Inter-class Specificity (orange) of ResNet50 and ViT on ImageNet-2012. The results show that the Fourier features identified by FFC exhibit higher intra-class concentration while maintaining inter-class specificity comparable to that of spatial-domain baselines. This indicates that the Fourier features discovered by FFC possess stronger structural characteristics than spatial features.
    (c) Minimum Features Required to Preserve Decisions. The x-axis represents the feature deletion rate, and the y-axis indicates the proportion of samples whose decisions are preserved. (d) The rectification rate of the different baselines after deleting the top-scored features.}
    \label{fig:structure}
\end{figure*}
\textbf{Intra-class Concentration}:
Kurtosis is used to measure the concentration, which reflects the ability of attribution methods to filter out noise. The higher the kurtosis, the stronger the noise-filtering capability. Within each class, feature tags are added up across all samples with the same label along each dimension to compute an overall score. We then calculate the kurtosis of the high-score features. The results in Figure~\ref{fig:structureRes} and Figure~\ref{fig:structureViT} indicate that Fourier features exhibit a higher concentration, with both kurtosis values approaching $1.8$ for ResNet and ViT-B/32. In contrast, spatial attribution methods exhibit lower concentration, barely exceeding random attribution by $0.2$ at best. \textbf{Thus key Fourier features are more likely to appear in the fixed positions.}\\
\textbf{Inter-class Specificity}: 
To assess inter-class specificity, the mean appearance times of high-score features across different classes are computed. Given the binarization of feature tags to $1$ for high and $0$ for low, a feature’s inter-class mean value approaching $1/1,000$ indicates greater specificity, given that $1/1000$ represents the feature is only present in one class. We count the number of features with an inter-class mean equal to $1/1000$, as a higher count suggests higher specificity. Figure~\ref{fig:structure} shows that while FFC's kurtosis is significantly higher than any baselines, its inter-class specificity metric remains stable above $3000$ in both models, ranking in the top $3$ among all methods.

Even though the features tend to appear in more concentrated and fixed regions, their specificity remains comparable to that of the spatial baselines. \textbf{This indicates that the features identified by FFC exhibit structural characteristics, supporting our claim.}
\\
\textbf{Minimum Features Required to Preserve Decisions}:
Although the confidence of the original category increases with the increase in the feature deletion rate, we find that the confidence of some other categories also increases. This experiment illustrates the minimum number of features required to maintain the original decision of the samples. We define maintaining the decision as the highest-confidence class of the processed sample being consistent with that of the original sample. As shown in Figure~\ref{fig:minimum}, for both backbone networks, only a small number of Fourier features are needed to maintain the original decision. For the ViT model, just $4\%$ of the Fourier features are sufficient to maintain the original decision for $80\%$ of the samples, while ResNet requires slightly more, at $15\%$ of the features to maintain the original decision for $80\%$ of the samples.

\textbf{Aligned with their structural properties, Fourier features provide a more suitable basis for explainable AI in classification.}

\subsection{FFC Supports Diverse Modalities and Models}
\begin{figure*}
    \centering
    \subfloat[]{
    \includegraphics[width=0.31\linewidth]{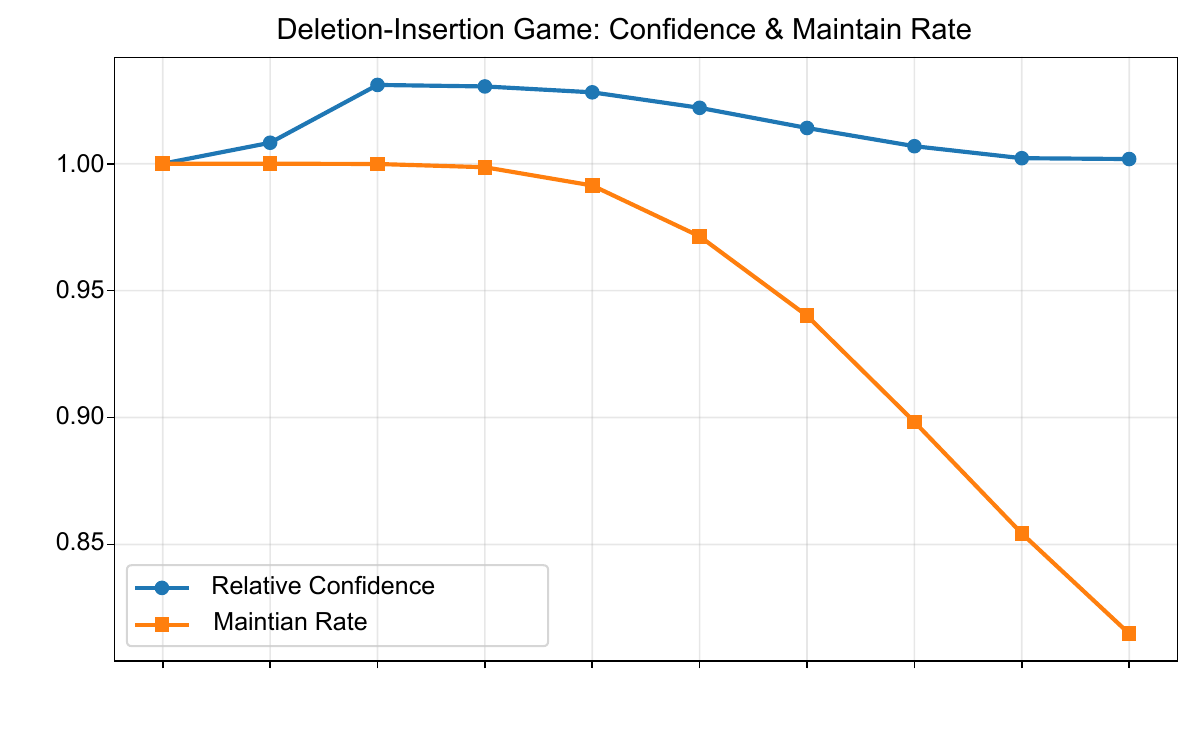}
    \label{fig:geneformer}
    }
    \hfil
    \subfloat[]{
    \includegraphics[width=0.31\linewidth]{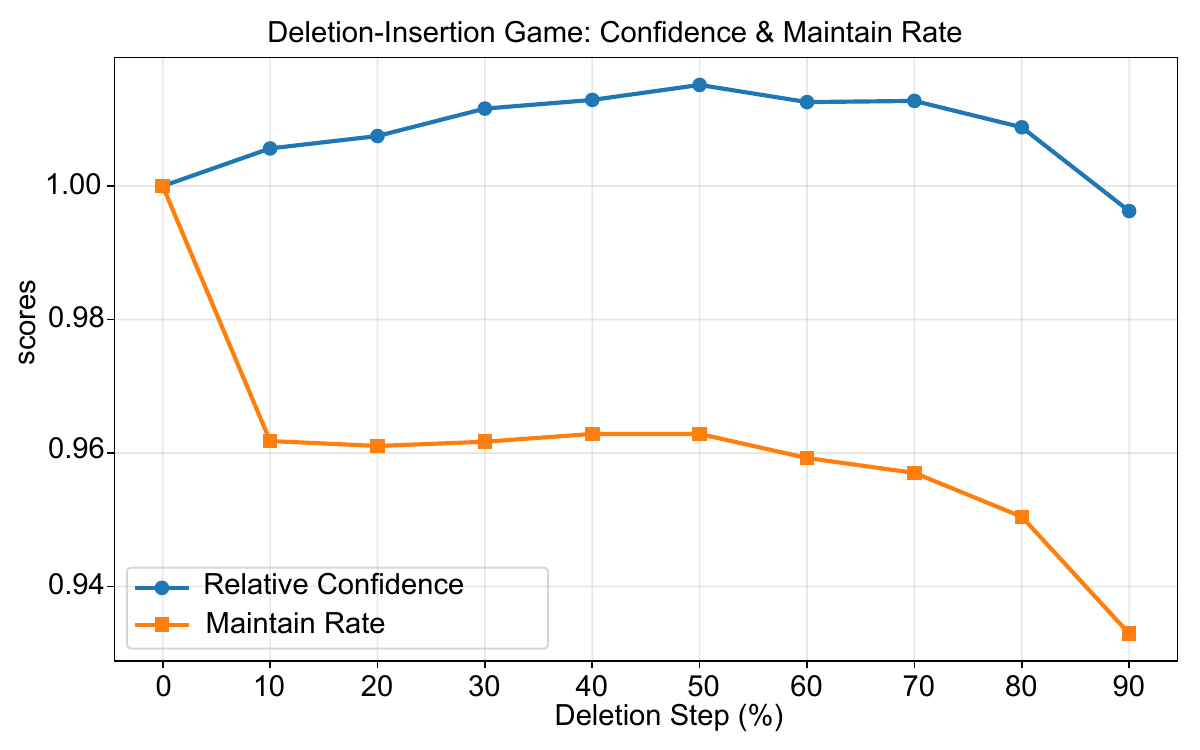}
    \label{fig:audio}
    }
    \hfil
    \subfloat[]{
    \includegraphics[width=0.31\linewidth]{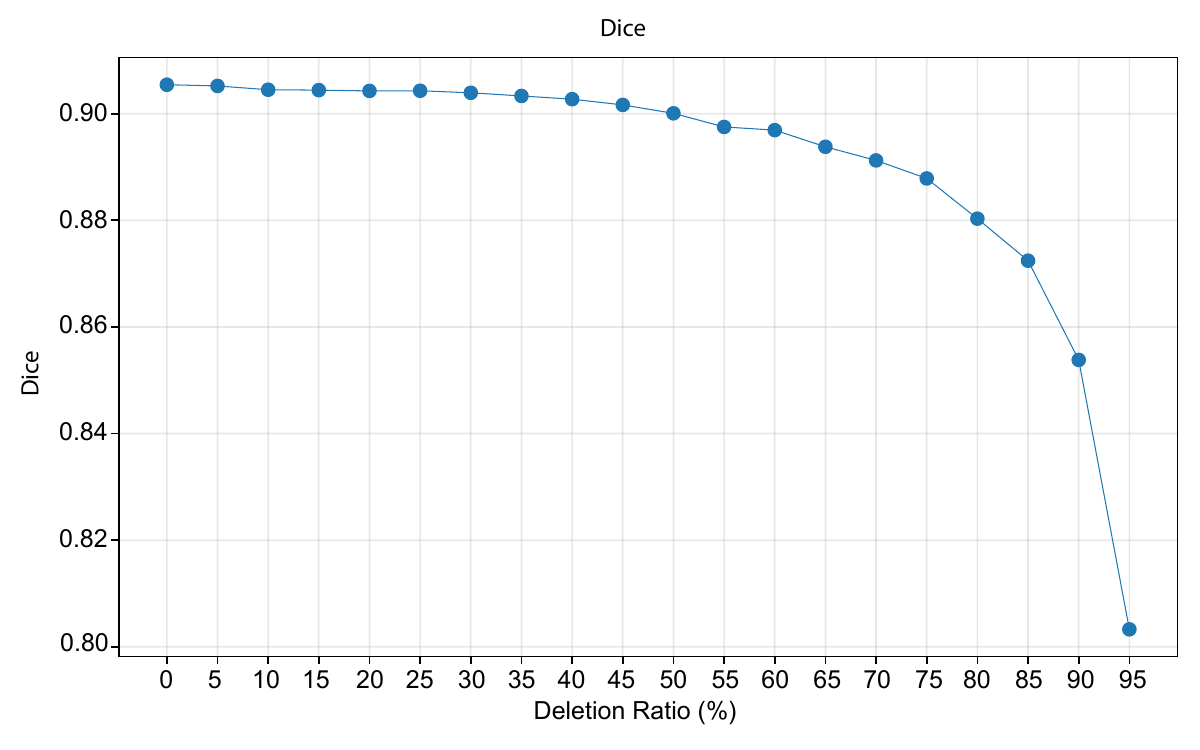}
    \label{fig:unet}
    }
    \caption{(a-b) The deletion-insertion game on multiple models (a: text, Geneformer; b: audio, ResNet). The x-axis represents the feature deletion rate, and the y-axis indicates the value of each point. The blue curve represents the relative confidence, and the yellow curve represents the maintain rate. 
    (c)The deletion-insertion game on UNet, segmentation task. The x-axis also represents the feature deletion rate. The y-axis indicates the dice value between original prediction and the prediction after deleting the features.}
    \label{fig:diversity}
\end{figure*}
Unlike traditional feature attribution, which mostly focuses on image and classification models, FFC supports multiple modern tasks and structures. We conduct our experiments on: \\
\textbf{1) Geneformer}~\cite{geneformer}: a large language model specialized in single-cell research. The model takes gene tokens as inputs and encodes them to latent-space. We use pre-trained Geneformer to train a cardiac disease prediction model. We apply FFC to the input embeddings directly. As shown in Figure~\ref{fig:geneformer}, with a clearly defined non-effect baseline of Fourier feature, the deletion-insertion game reliably shows that FFC can increase the relative confidence as well. Notably, the model also needs only a small number of Fourier features to maintain its prediction.\\
\textbf{2) ResNet-Audio}~\cite{heartmurmur}: We trained a ResNet-50 for heart murmur prediction. As shown in Figure~\ref{fig:audio}, FFC also increases the relative confidence. Additionally, the model needs only a small number of features for prediction.\\
\textbf{3) UNet}: UNet is trained on its provided dataset~\cite{Unet} in its original paper. The Dice coefficient, also known as the Dice similarity index, quantifies the overlap between the predicted segmentation and the ground-truth mask. It is computed as the formula below:
\begin{equation}
    Dice = \frac{2|P\cap G|}{|P|+|G|},
\end{equation}

\begin{figure}
    \centering
    \includegraphics[width=0.8\linewidth]{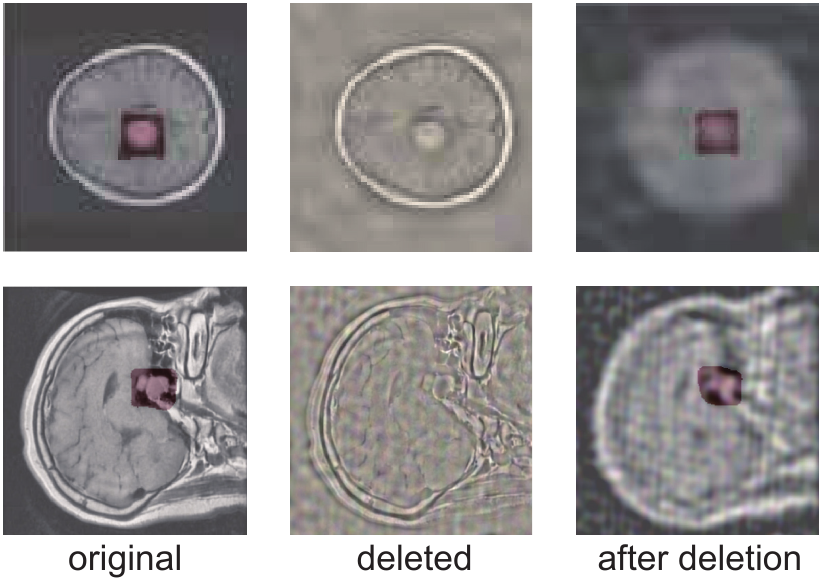}
    \caption{After deleting up to $95\%$ of the features, UNet can still recognize the objects in the MRI, even the deleted features can point out the tumor.}
    \label{fig:unet-vis}
\end{figure}
where $P$ denotes the set of modified positive pixels and $G$ denotes the original positive pixels. The numerator measures twice the intersection between the two sets, while the denominator reflects their combined size. A Dice score of 1 indicates perfect overlap, whereas a score of 0 indicates no overlap. As shown in Figure~\ref{fig:unet}, the Dice coefficient remains largely high even $80\%$ of the features are deleted. It is worth mentioning that, the segmentation task already highlights the important region, thus the traditional spatial attribution methods are redundant. However, FFC can reveal the most important signals rather than a region. We show the MRI after deleting up to $95\%$ of the features in Figure~\ref{fig:unet-vis} for directly visualization.

\subsection{Intuitive understanding of Fourier features}
\textbf{Visualizing the Comprehensive Information of Fourier Features by Rectifying Errors}:
To visually understand the semantic meanings of the results of Fourier features attribution, we select Fourier features that cause misclassifications and transform them back to the spatial domain for interpretation. Specifically, using the FFC, we identify high-score features in misclassified samples (approximately $20\%$ of the samples) and sequentially remove the top-scoring features (we assume they are the most important features that cause misclassifications), ranging from $1/10,000$ to $1/1,000$, with a step size of $1/10,000$. The process stops when the sample's classification is corrected (i.e., the highest-confidence class matches the true label). A correction is considered failed if the sample cannot be corrected after removing the top $1/1,000$ of the features. We systematically compare the correction rates of FFC to those of the baseline methods (random removal, removal based on frequency magnitude, and removal based on energy magnitude). Figure~\ref{fig:cor-rate} shows that the correction rate of FFC ranges from $25\%$ to $28\%$, which is significantly higher than the random removal($3\%$) and the frequency magnitude-based removal($20\%$-$22\%$). However, there is no significant difference compared to the energy-based removal method. Due to simply removing the top features instead of finding all the features causing the misclassification, FFC still corrects errors to a certain extent and holds considerable potential for future practical applications.

\begin{figure}
    \includegraphics[width=0.9\linewidth]{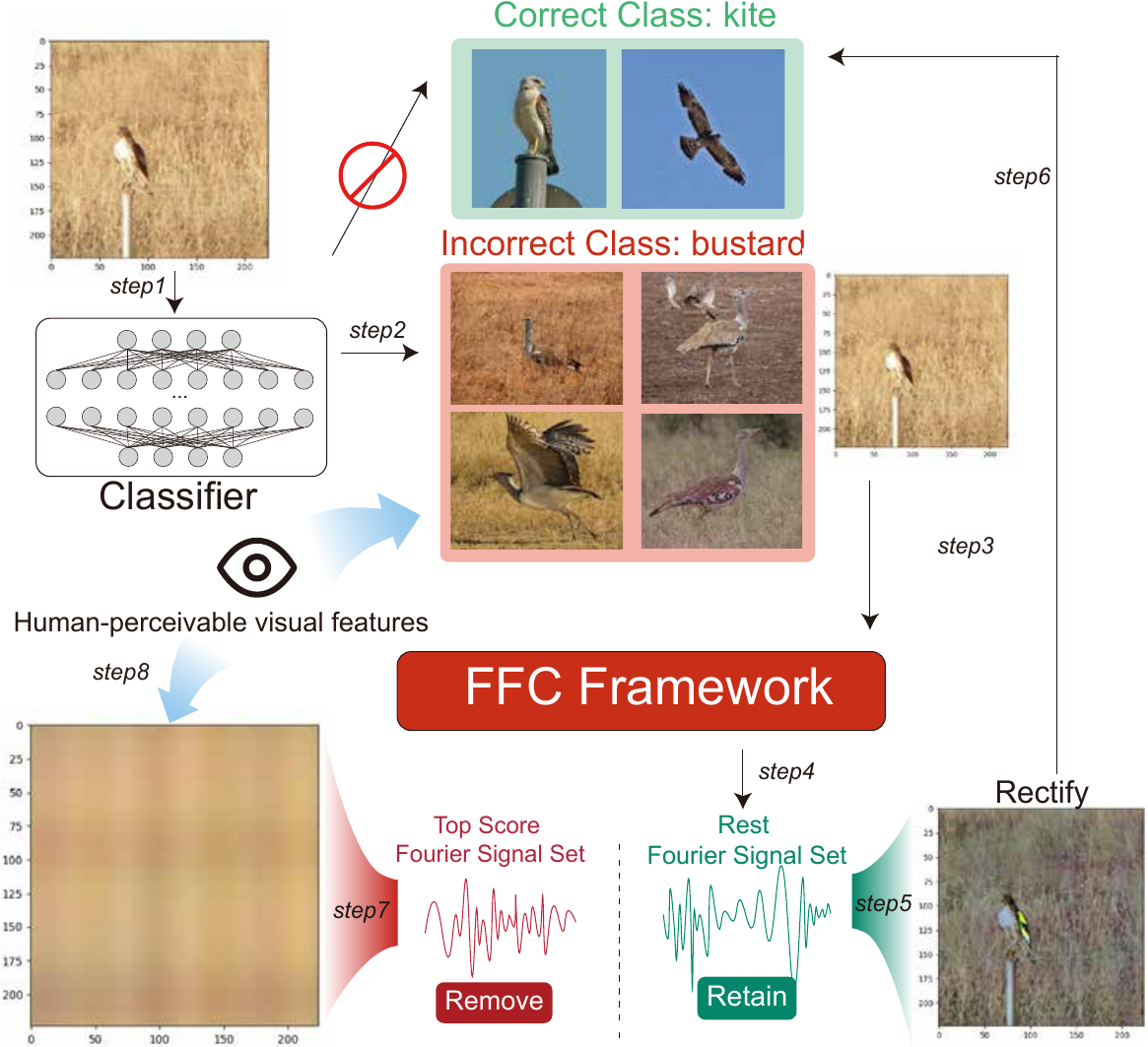}
    \label{fig:cor-proce}
    \caption{Understanding the Semantic Meaning Carried by Fourier Features. Compared to spatial-domain attribution, Fourier feature attribution can reveal the semantic meaning of features, whereas spatial attribution typically only highlights regions—and these regions may even be misleading. The schematic pipeline of the procedure is as follows:a sample is first input to the network, which initially misclassifies it as the category bustard. Using FFC, the importance of each feature is ranked, and the top $1/10,000$ features are removed. The rectified sample is then visualized, after which the network correctly classifies it as kite. The deleted features are also visualized, and their correlation with the original misclassified category (bustard) is analyzed. Notably, the deleted features resemble the yellow background characteristic of the bustard category, demonstrating that Fourier feature attribution captures meaningful semantic information beyond merely identifying spatial regions.}
    \label{fig:fouriermeaning}
\end{figure}

Figure~\ref{fig:cor-proce} and Figure~\ref{fig:intro-b} illustrate an example from the dataset, the $42$-th image of class $21$ kite. This image is misclassified by the ViT to the class of $138$, bustard. The image is selected as an example because after removing only $15$ (approximately $1/10,000$) high-score features, it is corrected to its true class, kite, suggesting the high contribution of the $15$ high-score features to the cause of the misclassification. We transform the removed Fourier features back to the spatial domain and observe that the deleted signals, which cause the misclassification, exhibit visual similarities to class bustard (the original misclassified class). In this example, both the deleted features and the samples of bustard display a yellow background. In contrast, most samples in class kite have a blue sky background, while class bustard contains more samples with yellow backgrounds (Figure~\ref{fig:cor-proce}). Based on this, we infer that the network misclassifies the sample as it incorrectly extracts the feature of the yellow background and associates it with the class bustard, which genuinely has yellow background features. This means that the network might mistakenly take the background information as a feature of the bustard. Using FFC, we successfully and interpretably correct the network's decision. 

Compared with the example in Figure~\ref{fig:intro-b}, \textbf{Fourier feature attribution can reveal richer semantic information than spatial attribution, even when using only $15$ Fourier features.} By contrast, $15$ pixel features cannot reveal such rich semantic information.

\section{Conclusion}
This paper demonstrates the rationale for applying game-theory-based metrics in the context of Fourier feature attribution and introduces the Fast Fourier Correlation (FFC) algorithm, grounded in control theory, offering a novel perspective for frequency-domain feature attribution. Experimental results validate the effectiveness of FFC, highlighting its superior feature selection capabilities and distinguishing it from any existing methods. Comparisons between spatial- and frequency-domain attribution reveal that Fourier features exhibit stronger intra-class concentration while maintaining comparable inter-class specificity, suggesting that they are particularly well-suited for classification tasks and the development of explainable AI. The manually constructed dataset and adversarial experiments further confirm the faithfulness of FFC. Extensive evaluations on Geneformer, UNet, and audio datasets demonstrate the generalizability of the proposed method. Additionally, visualization of Fourier features through misclassification correction analysis shows that the corresponding spatial features are perceptible to humans.

This work does not systematically explore the relationship between the Fourier features identified by FFC and the underlying causes of misclassification. Future studies could address this gap, providing deeper insights into interpretable AI. Moreover, related investigations could have significant implications for fields such as backdoor detection and adversarial sample generation. Building on FFC, further research may advance the development of explainable AI, enhancing the reliability and robustness of AI systems and addressing limitations of current attribution methods. 

\bibliography{ref}
\bibliographystyle{IEEEtran}

\vfill

\end{document}